\UseRawInputEncoding
\documentclass[lettersize,journal]{IEEEtran}
\usepackage{amsmath,amsfonts}
\usepackage{algorithmic}
\usepackage{array}
\usepackage[caption=false,font=normalsize,labelfont=sf,textfont=sf]{subfig}
\usepackage{textcomp}
\usepackage{stfloats}
\usepackage{url}
\usepackage{verbatim}
\usepackage{graphicx}

\usepackage{booktabs}
\usepackage{multirow}
\usepackage{tabu}
\usepackage{graphicx}
\usepackage{array}
\usepackage{colortbl}

\def\BibTeX{{\rm B\kern-.05em{\sc i\kern-.025em b}\kern-.08em
    T\kern-.1667em\lower.7ex\hbox{E}\kern-.125emX}}
\usepackage{balance}
\begin{document}
\title{Semantic decomposition Network with Contrastive and Structural Constraints for Dental Plaque Segmentation}
\author{Jian Shi, Baoli Sun, Xinchen Ye, Zhihui Wang, Xiaolong Luo, Jin Liu, Heli Gao, Haojie Li
\thanks{This work was supported by National Natural Science Foundation of China (NSFC) under Grant 61702078, 61772108, 61976038, 61772106.}
\thanks{Jian Shi, Baoli Sun, Xinchen Ye, Zhihui Wang, Haojie Li are with DUT-RU International School of Information Science \& Engineering, Dalian University of Technology, Liaoning, and Key Laboratory for Ubiquitous Network and Service Software of Liaoning Province, China (Corresponding author: Zhihui Wang. E-mail: zhwang@dlut.edu.cn).}}

\markboth{}%
{How to Use the IEEEtran \LaTeX \ Templates}

\maketitle

\begin{abstract}
Segmenting dental plaque from images of medical reagent staining provides valuable information for diagnosis and the determination of follow-up treatment plan. However, accurate dental plaque segmentation is a challenging task that requires identifying teeth and dental plaque subjected to semantic-blur regions (i.e., confused boundaries in border regions between teeth and dental plaque) and complex variations of instance shapes, which are not fully addressed by existing methods. Therefore, we propose a semantic decomposition network (SDNet) that introduces two single-task branches to separately address the segmentation of teeth and dental plaque and designs additional constraints to learn category-specific features for each branch, thus facilitating the semantic decomposition and improving the performance of dental plaque segmentation. Specifically, SDNet learns two separate segmentation branches for teeth and dental plaque in a divide-and-conquer manner to decouple the entangled relation between them. Each branch that specifies a category tends to yield accurate segmentation. To help these two branches better focus on category-specific features, two constraint modules are further proposed: 1) contrastive constraint module (CCM) to learn discriminative feature representations by maximizing the distance between different category representations, so as to reduce the negative impact of semantic-blur regions on feature extraction; 2) structural constraint module (SCM) to provide complete structural information for dental plaque of various shapes by the supervision of an boundary-aware geometric constraint. Besides, we construct a large-scale open-source Stained Dental Plaque Segmentation dataset (\textit{SDPSeg}), which provides high-quality annotations for teeth and dental plaque. Experimental results on \textit{SDPSeg} datasets show SDNet achieves state-of-the-art performance. The dataset is in https://github.com/anaanaa/SDPSeg-dataset

\end{abstract}
	%
	%
\section{Introduction}
\label{sec:intro}
Dental plaque is a soft, unmineralized  biofilm strongly adhering to tooth surface which leads to carious lesions such as dental caries and periodontitis
\cite{b1, b2, b3}.
Plaque stain reagent is commonly used in the clinical practice to stain dental plaque and the Quigley-Hein modified by Turesky DP index (Q-H/TPI) \cite{b4, b5} is used to assess the oral hygiene state by scoring the distribution of dental plaque.
However, estimation of the distribution of dental plaque
depends on the clinician's experience, leads to a high degree of subjectivity.
The accuracy of evaluation varies dramatically among different physicians which drives us to using computer-aided diagnosis (CAD) systems for automatic diagnosis of dental plaque to get a more objective evaluation result.
Segmentation of dental plaque is a fundamental step to build such CAD systems and crucial for accurate quantification of dental plaque  distribution.

\begin{figure*}[ht]
	\centering
	\centerline{\includegraphics[width=1\linewidth]{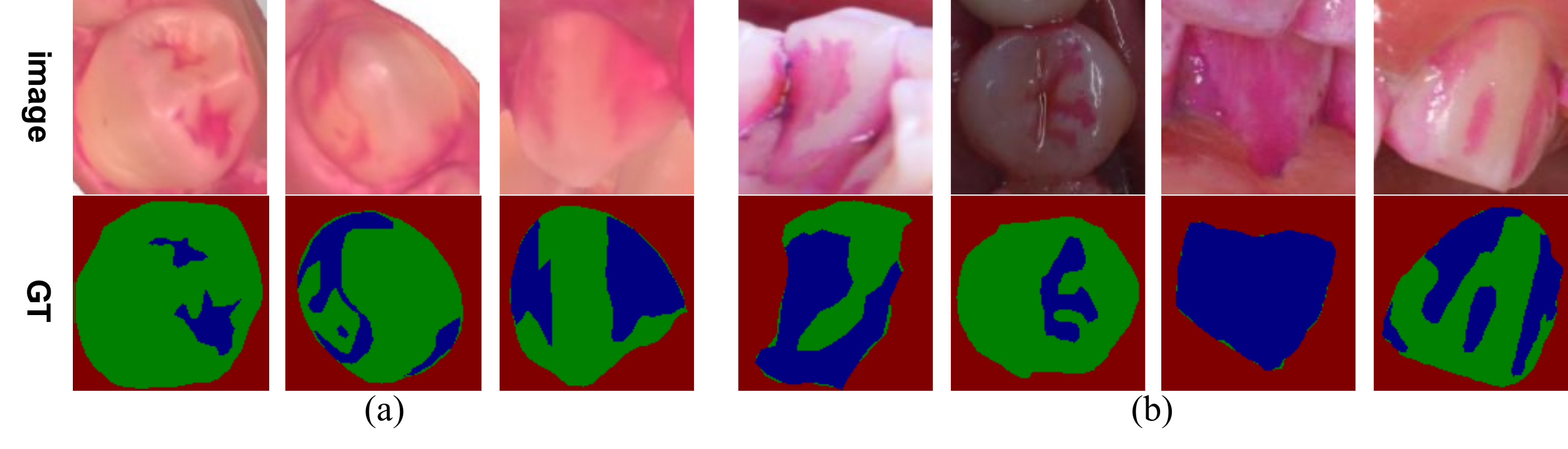}}
	\vspace{-8pt}
	\caption{Examples of oral images and corresponding labels in our proposed \textit{SDPSeg} dataset. (a) and (b) show examples obtained by high-resolution laser scanner (\textit{SDPSeg-S}) and high-definition camera (\textit{SDPSeg-C}), respectively. Two major challenges in dental plaque segmentation can be observed from these examples: the varied shapes and sizes of dental plaque and the blurred boundaries between teeth and dental plaque.}
	\label{fig:intro}
	\vspace{-6pt}
\end{figure*}

Recently, there have been some works on the classification and segmentation of dental plaque. \cite{b6} presented an automated dental plaque image classification model based on Convolutional Neural Networks (CNN) on Quantitative Light-induced Fluorescence images. \cite{b7} conducted low-shot learning at the super-pixel level of automatic dental plaque segmentation based on local-to-global feature fusion. Although existing dental plaque segmentation method has demonstrated remarkable progress, the discriminative ability in semantic-blur regions (i.e., confused boundaries in border regions between teeth and dental plaque)  still remained to be poor, as \cite{b7} neglected to distinguish the contextual dependencies of different categories, which may result in a less reliable context especially in semantic-blur regions.
There are still several limitations in accurate segmentation of teeth and dental plaque:
(1) as Fig. \ref{fig:intro} shown, the semantic-blur regions with low contrast between dental plaque and teeth have lower confidence in classification. Therefore, how to reduce the negative impact of semantic-blur regions on feature extraction of teeth and dental plaque still needs to be studied;
(2) complex structure information caused by high frequency variations in shape and size of the dental plaque instances increases the difficulty of segmentation.

\begin{figure*}[ht]
	\centering
	\centerline{\includegraphics[width=1\linewidth]{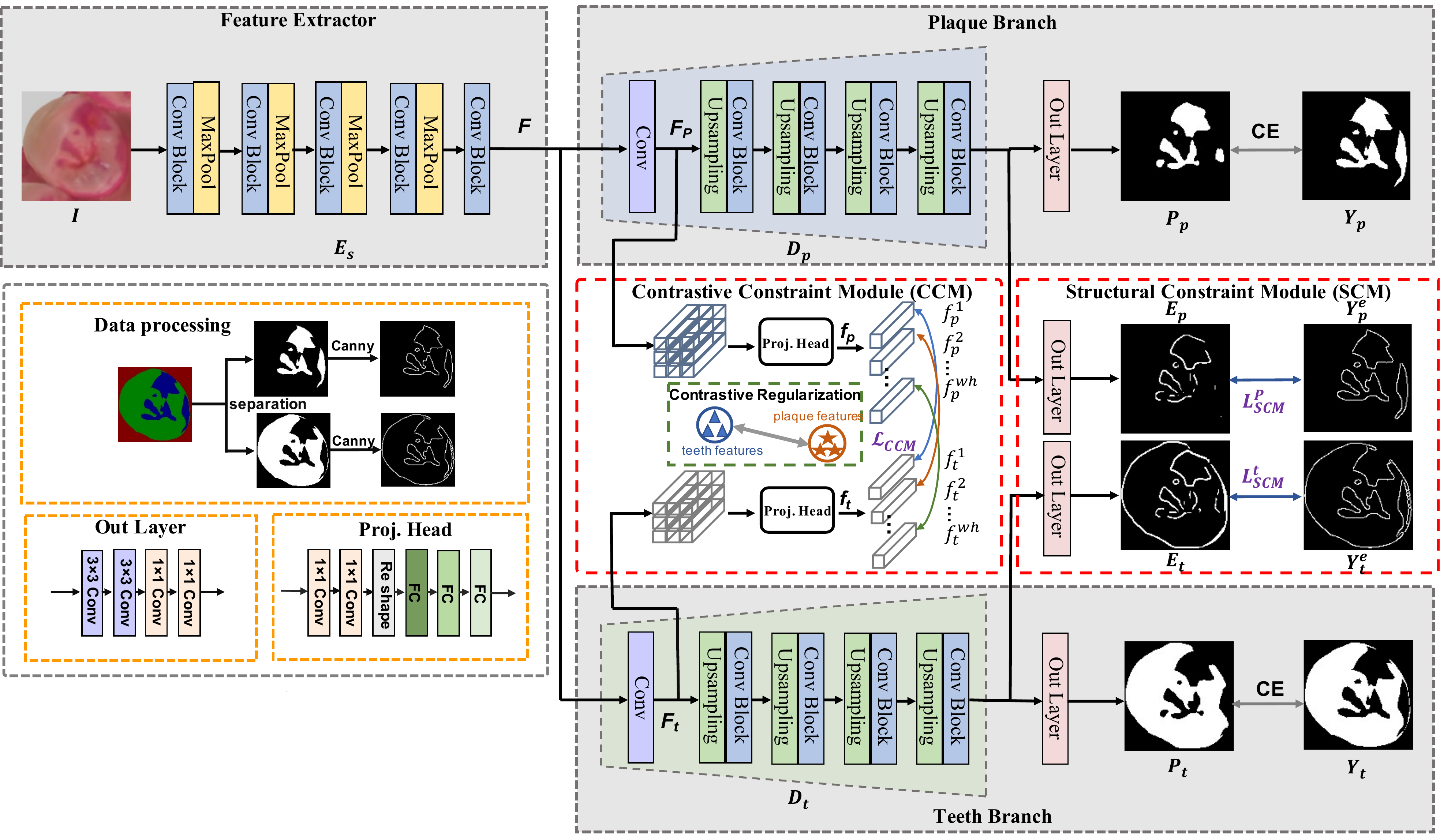}}
	\vspace{-6pt}
	\caption{The Overall network architecture. The "feature extractor" is shared by the decoupled two-branch subnetworks which produce dental plaque prediction and teeth prediction in plaque branch and teeth branch, respectively. Contrastive constraint module (CCM) maximize the distance between category representations in the semantic-blur regions. Structure constraint module (SCM) improves the structural integrity of instance by the supervision of an boundary-aware geometric constraint.}
	\label{fig:mainnet}
	\vspace{-6pt}
\end{figure*}

Motivated by the above analysis, this paper introduces a novel semantic decomposition network (SDNet) which employs divide-and-conquer strategy for the task of dental plaque segmentation, as shown in Fig. \ref{fig:mainnet}. 
We explore for the first time to disentangle the dental plaque segmentation into two sub-tasks, i.e., teeth segmentation and dental plaque segmentation, through two independent decoder branches. The decomposed branches decouple the entangled  relation between teeth and dental plaque, making the network only focus on each category and distinguish semantic-blur regions through clearer supervision information for each branch.

In order to learn category-specific feature representations for two separated branches, we introduce additional constraints from the space of latent features to clearly distinguish instances of different categories.
Motivated by our observation that the regions around the interaction of teeth and dental plaque are more category-uncertainty than others, we devise a contrastive constraint module (CCM) based on contrastive learning \cite{b8, b9, b10}
to enlarge the distance between the teeth and dental plaque category in the latent space, especially in the semantic-blur regions, which provides a more discriminative feature representation for semantic information decomposition.
Secondly, to reduce the structural uncertainty caused by irregular variations of shape and size in different dental plaque instances, we further propose a structural constraint module (SCM) that provides holistic structure information for dental plaque by an additional pixel-level supervision of a boundary-aware geometric constraint, so as to guide the learning of spatial details especially in the semantic-blur regions.  

Besides, deep learning methods of semantic segmentation achieve good performance by training deep models with massive labeled data, but they still suffer from unreliable adaptation and overfitting when training on a small-scale dataset.
As a result, a large-scale and high-quality dental plaque segmentation dataset is needed for the development of automatic diagnosis of dental plaque.
However, stained dental plaque data is rare as the collection limited by the privacy of patients, and the large quantities of annotated data is also expensive. To the best of our knowledge, there is no specially designed stained dental plaque public dataset for CAD,  which impedes the development of this research area.
Therefore, we  construct  a  large-scale  open-source  Stained Dental Plaque Segmentation dataset (\textit{SDPSeg}) for research and assessment of the task.
In \textit{SDPSeg}, we present two high quality public datasets, \textit{SDPSeg-S} and \textit{SDPSeg-C}  which contain 565 and 1304 images captured by high-resolution laser scanner and HD camera, respectively.
Both datasets contain three categories: teeth, dental plaque and background, each image contains a teeth instance and zero or several dental plaque instances.
The labels in two datasets are meticulously annotated at pixel level by experienced experts, fine grained dental plaque and complex contours are also annotated in detail.

The key contributions are summarized as follows:

$\bullet$  We propose a novel semantic decomposition network that employs a divide-and-conquer strategy to effectively segment the dental plaque on teeth with two decomposed branches.

$\bullet$  We further devise two constraint modules (CCM and SCM) to address the problem of semantic-blurs and instance shape variance by explicitly modeling the dissimilarity of the features of different categories through contrastive learning and providing holistic structural information for dental plaque with variant shapes and sizes through a boundary-aware geometric constraint.

$\bullet$ We develop two large-scale datasets (\textit{SDPSeg-S},  \textit{SDPSeg-C}) that cover a wide range of age groups and conform to the real situation of dental plaque distribution in clinical diagnosis, which is suitable for the real-world CAD of dental plaque.

$\bullet$ Extensive experiments on our proposed \textit{SDPSeg} datasets show that our SDNet outperforms previous state-of-the-art methods without introducing additional computation during inference time, since both additional constraints are only required in training stage. Our method is more accurate in estimating the ratio of dental plaque to teeth compared to clinicians, which indicates that our effectiveness in CAD.

\section{Related Work}

\textbf{Medical image segmentation.}
Most semantic segmentation frameworks based on convolutional neural networks are either full convolutional networks (FCN) \cite{b11} or U-shaped networks based on an encoder-decoder architecture like UNet \cite{b12} designed for HeLa cells and neuronal structures of electron microscopic stacks. Various modifications based on these two forms have been proposed for natural image segmentation
\cite{b16, b17}
 and medical image segmentation 
 \cite{b23, b24}. 
 The purpose of polyp segmentation is to accurately segment polyps from a given colonoscopy image. UACANet \cite{b25} proposed uncertainty augmented context attention network to augment the uncertainty of the boundary area. Parallel reverse attention network \cite{b26} utilizes a parallel partial decoder to generate the high-level semantic global map and three reverse attention modules to mine complementary regions and details from high-level features. In the field of retinal vessel segmentation, the semantic aggregation module and multi-scale aggregation module were used in \cite{b27} to obtain more semantic feature representation and extract multi-scale information,  respectively. 
 CAG-Net \cite{b28} employs prediction module and refinement module to generate more accurate results for retinal vessel segmentation. A Study Group Learning (SGL) scheme \cite{b29} was proposed to improve the robustness of the model which trained on noisy labels. 
 In 3D prostate segmentation, MS-Net \cite{b30} utilized a multi-site network which learned robust representation and leveraged multiple sources of data for improving segmentation results. 
 Jia \textit{et al.} \cite{b31} proposed a 3D adversarial pyramid convolutional deep neural network which consisted of a generator performing image segmentation and a discriminator that distinguished segmentation result and its corresponding groundtruth. Medical image segmentation plays a great role in the process of CAD. However, in the clinical diagnosis of dental plaque, the application of CAD is still insufficient.

\textbf{Dental plaque segmentation.}
Dental plaque segmentation task aims to accurately segment the dental plaque area and teeth area with medical reagent staining.
In this paper, we first propose a stained dental plaque segmentation task and an open-source stained dental plaque segmentation dataset. In other tasks related to dental plaque, such as dental plaque classification task and dental plaque segmentation, \cite{b6} proposed a dental plaque image classification dataset containing 427 images obtained by Quanti tative Light-induced Fluorescence (QLF) camera where dental plaque fluoresces red, and presented an automated dental plaque image classification model based on Convolutional Neural Networks (CNN). Li \textit{et al.} \cite{b7} proposed a dental plaque segmentation dataset containing 607 images without staining obtained by oral endoscopes, the dataset is private, and the annotation method that dentists mark the regions of plaque referring to the post-stained images leads to inaccurate annotation. \cite{b7} also proposed an automatic dental plaque segmentation network which fuse multi-scale contextual features using traditional algorithms like heat kernel signature (HKS) \cite{b32} and local
binary pattern (LBP) \cite{b33} for the daily monitoring of the patient with no reagent staining.
Unfortunately, there is a lack of a public dataset of stained dental plaque and research studies in the clinical diagnostic application of dental plaque segmentation. Therefore, we present an open-source dental plaque staining dataset and a semantic decomposition framework utilizing divide-and-conquer strategy and two constraint modules which are contrastive constraint module (CCM) and structure constraint module (SCM) to facilitate the semantic decomposition and improve the performance of dental plaque segmentation.

\textbf{Divide-and-conquer strategy.} Lin \textit{et al.} \cite{b34}, Kim \textit{et al.} \cite{b35}, Huang \textit{et al.} \cite{b36} and Xu \textit{et al.} \cite{b37} proposed adversarial learning network based on divide-and-conquer strategy for generation, super-resolution, enhancement of images and robust semantic segmentation, respectively. Lin \textit{et al.} \cite{b34} reduced memory requisition during training and inference by dividing image generation into separated parallel sub-procedures. Kim \textit{et al.} \cite{b35} designed a novel GAN-based joint SR-ITM network with a divide-and-conquer approach which was divided into three task-specific subnets: an image reconstruction subnet, a detail restoration (DR) subnet and a local contrast enhancement (LCE) subnet. Huang \textit{et al.} \cite{b36} decomposed the photo enhancement process at three levels of perception, frequency and dimension. Xu \textit{et al.} \cite{b37} employed dynamical division mechanism and divided pixels into multiple branches automatically. In the field of medical image segmentation, we innovatively utilize the divide-and-conquer strategy to divide the dental plaque segmentation task into two sub-problems, which uses two category-specific branches to solve the segmentation of teeth and dental plaque, respectively.

\textbf{Contrastive learning.} Contrastive learning \cite{b38, b39, b40}, a successful variant of self-supervised learning (SSL), forcing the embedding features of similar images to be close in the latent space and those of dissimilar ones to be apart, is often used in semi-supervised or self supervised medical image segmentation with limited annotations. Krishna \textit{et al.} \cite{b41} proposed a novel domain-specific contrastive strategies leveraging the structural similarity across volumetric medical images. Hu \textit{et al.} \cite{b42} proposed a supervised local contrastive loss which forced pixels with the same label to gather in the embedded space with limited pixel-level annotations. Bai \textit{et al.} \cite{b43} introduced contrastive learning to enhance contextual relationships between pixels in a dataset rather than just in an image.
However, in this paper, we employ CCM which employs contrastive loss to construct positive and negative pairs of teeth and dental plaque category in an image to enlarge the differences between different categories as an dditional constraint in the supervised learning.

\textbf{Boundary aware information.} The boundary of object is often used in medical image segmentation as guidance information. Wang \textit{et al.} \cite{b57}proposed a novel boundary coding network (BCnet) to learn a discriminative representation for organ boundary and use it as the context information to guide the segmentation. Wen \textit{et al.} \cite{b58} presented two knowledge distillation modules, namely boundary-guided deep supervision and 
output space boundary embedding alignment, to explicitly transfer boundary information. Zhang \textit{et al.} \cite{b59} and \textit{et al.} \cite{b60} shared a similar boundary attention idea, where the object boundary is implicitly extracted from region predictions with a foreground erasing mechanism. Different from the above methods , we integrate the boundary-aware geometry constraint CCM into the loss function to provide holistic structure information for dental plaque.

\section{Proposed Method}\label{PDPA}

\subsection{Overview}
	
The overview of our proposed semantic decomposition network (SDNet) is illustrated in Fig. \ref{fig:mainnet}. The whole network is mainly divided into three components: (1) a shared feature encoder for spatial feature extraction;
(2) two  category-specific segmentation branches, i.e., dental plaque branch and teeth branch, for predicting teeth and dental plaque segmentation masks, respectively; (3) two constraint modules consisting of contrastive constraint module (CCM) and structure constraint module (SCM), for introducing additional constraints from the space of latent feature to clearly distinguish instances of different categories in the semantic-blur regions between teeth and dental plaque and provide complete structural
information for dental plaque with various shapes.

The goal of our SDNet is to improve the segmentation accuracy of dental plaque by decoupling the entangled relation between dental plaque and teeth in a divide-and-conquer strategy. Given an input image $I$, the shared feature encoder first extracts the intermediate features $F$ for latter segmentation branches through continuous convolution layers and downsampling operations. Then the prediction is generated by decoding the required category-specific features from the shared encoder through different supervisions in each branch. CCM is introduced in the front stage of the decoder that can model the irrelevance of the category-specific features in high level, which enlarges the difference between categories and obtain more discriminant features for decomposition.
Structure constraint module (SCM) is introduced in the last stage of the decoder that can improve the structural integrity of instances by introducing low-level features to recover the boundary detailed information.

\subsection{Shared feature encoder }\label{RB}
As shown in Fig. \ref{fig:mainnet}, we employ a shared encoder $E_s$ to extract both teeth and dental plaque features from an input oral image $I$. The encoder network $E_s$ contains 5 stacked convolution blocks.
Each convolution block consists of the typical architecture of two $3\times3$ convolution layers, a rectified linear unit (ReLU) activation function, and each convolution block is followed by a $2\times2$ max pooling operation with stride 2 for feature downsampling except for the last one.

\subsection{Semantic Decomposition}\label{semantic decomposition}
 As we observed that many segmentation errors generally occur in the conflict areas between categories, especially the junction areas or overlapping areas. Previous approaches resolved the prediction conflicts between neighboring objects through context prior information or additional post-processing \cite{b44}.
 Consequently, their results are over-smooth along boundaries or exhibit small gaps between neighboring objects.
 Motivated by the divide-and-conquer strategy, we design two independent branches to decouple the  entangled relation of teeth and dental plaque, which employ unique feature decoders to decode category-specific features as required according to the difference of semantic information between categories.
 The different feature decoder modules aims to generate category-specific prediction for teeth and dental plaque with the guidance of corresponding supervision.

 Specifically, given the feature maps  $F$ from the shard encoder $E_s$,
 the dental plaque decoder $D_p$ and teeth decoder $D_t$ produce dental plaque prediction and teeth prediction, respectively. The branch for dental plaque prediction is designed in a simple yet effective way: a decoder with four stacked blocks
 where each block contains a upsampling layer and a convolution block. And finally, an output layer with a series of convolution layers is applied to obtain mask predictions. The structure of teeth prediction branch is the same as that of dental plaque prediction. The segmentation loss $L_S^{p}$ for the dental plaque mask prediction is defined as:
\begin{align}
& L_{S}^{p} = L_{CE}(P_{p},Y_{p}), \label{(2)}
\end{align}
and the loss for teeth mask prediction is defined as:
\begin{align}
& L_{S}^{t} = L_{CE}(P_{t},Y_{t}), \label{(3)}
\end{align}
where $L_{CE}$ denotes the cross-entropy loss function. $P_{p}$, $P_{t}$ denote the dental plaque prediction and teeth prediction respectively, and $Y_{p}$, $Y_{t}$  denote the binary groundtruths for the dental plaque and teeth obtained by channel separation from the original mask.

\subsection{Contrastive Constraint Module (CCM)}\label{CCM}

Contrastive Learning aims to learn discriminative representations by comparing positive and negative pairs which is usually used in the field of self-supervised representation learning to keep the distribution of unlabeled data consistent with that of labeled data.
Specifically, our key idea is to employ contrastive learning innovatively to maximize the divergence among different categories and strengthen category-specific features in the supervised learning.

As shown in Fig \ref{fig:mainnet}, we introduce CCM before the first layer of each separated decoder to push two representations away from each other in high-level feature space.
Given two feature maps $F_{p}$, $F_{t}$ with the spatial resolution $w\times h$, and $C$ feature channels from the convolution layer before the first stage of dental plaque decoder $D_p$ and teeth decoder $D_t$, respectively, two projection head $ \textit{proj}_p$, $\textit{proj}_t$ are introduced to map two feature maps $F_{p}$ and $F_{t}$ into the vector spaces $\{f_p^i\}_{i=1}^{wh}$ and $\{f_t^i\}_{i=1}^{wh}$ where each pixel-level feature $f_p^i$ or $f_t^i$ is regarded as an independent class-wise representation for contrastive constraint. The  projection head can be instantiated as any of the existing projection heads such as the ones in \cite{b8, b39}, which takes the dense feature maps as input and generates multiple feature vectors for each class representation. In this paper, each projection head consists of a series of convolution layers used for reducing channel dimensions and three fully connected layers. The latent dental plaque feature representations are obtained by:
\begin{align}
& \{f_p^i\}_{i=1}^{wh} = \textit{proj}_p(F_p). \label{(7)}
\end{align}

Similarly, we obtain the teeth representations through $P_t$:
 \begin{align}
& \{f_t^i\}_{i=1}^{wh} = \textit{proj}_t(F_t). \label{(7)}
\end{align}

Then, the cosine similarity is used to measure the difference between dental plaque and teeth category-specific representations in latent feature space:
\begin{align}
& L_{CCM} = \sum_{i=1}^{wh} \frac{f_{p}^i}{\parallel
f_{p}^i \parallel}_2 \cdot {\frac{f_{t}^i}{\parallel f_{t}^i \parallel}_2}, \label{(7)}
\end{align}
where $||\cdot||_2$  is L2-normalized of corresponding feature representation. The contrastive constraint loss is devised based on constrastive learning \cite{b45}, and enforces the class-separability of pixel-level features between dental plaque and teeth branches.

\subsection{Structure Constraint Module (SCM)}\label{SCM}

For dental plaque with varied shapes and sizes, the complete structure of dental plaque instances is difficult to accurately predict. In order to get more structure information, we introduce the structure constraint module (SCM) which employs the supervision of an boundary-aware geometric constraint. Specifically, SCM is designed to predict boundary-aware mask model in both two branches.

\begin{figure}[ht]
	\centering
	\centerline{\includegraphics[width=0.95\linewidth]{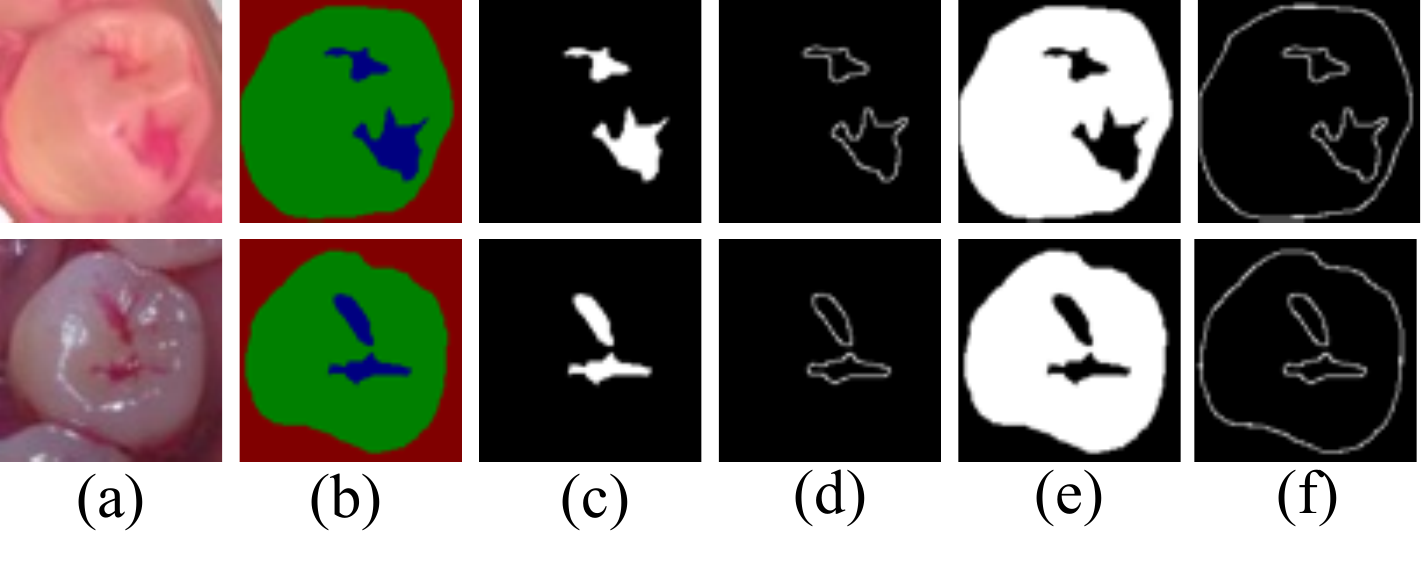}}
	\caption{The results of Canny operator. (a) is the original image, (b) is the Ground Truths, (c) is the mask Ground Truths of plaque through channel separation, (d) is the boundary Ground Truths of plaque through Canny algorithm, (e) is the mask Ground Truths of teeth through channel separation, (f) is the boundary Ground Truths of teeth through Canny algorithm.}
	\label{fig:canny}
    \end{figure}

We employ the canny operator (Fig \ref{fig:canny} shows the samples generated by canny operator, the first row is the results on \textit{SDPSeg-S} dataset, and the second row is the results on  \textit{SDPSeg-C} dataset) to generate the binary boundary maps $Y_p^e$ and $Y_t^e$ for dental plaque and teeth from the binary mask ground truths $Y_p$ and $Y_t$ .
Structure constraint module (SCM) updates the decoder parameters shared with mask predictions through back propagation, achieving the purpose of refining segmentation outputs. Therefore, it is particularly important to use an effective loss function which can constrain convolution network to generate an accurate and sharp boundary prediction.
Highly imbalanced categories of boundary versus background in training data leads that the boundary constraint is not robust enough. Followed \cite{b46}, we employ compound boundary loss that contains binary cross-entropy loss and dice loss \cite{b47} to optimize the imbalanced boundary prediction.

Dice loss only measures the distance between predictions and groundtruths at pixel level regardless of which category the pixel belongs to, thus reducing the imbalance between foreground and background pixels.
Our loss $L_{SCM}^{p}$ for dental plaque structure constraint is:
\begin{align}
& L_{SCM}^{p} = \alpha L_{BCE}(E_{p},Y_{p}^e) + \beta L_{Dice}(E_{p},Y_{p}^e),
\end{align}
where $E_{p}$ denotes the predicted boundary for  dental plaque and
$Y_{p}^e$ denotes the corresponding boundary groundtruth. $\alpha$, $\beta$ is a hyper parameter to adjust the weight of BCE loss and Dice loss. 
Dice loss is formulated as follows:
\begin{align}
& L_{Dice} = 1 - \frac{2 \sum_{i}^{H \times W} p_{p}^i y_{p}^i + \tau}{\sum_i^{H \times W}(p_{p}^i)^2 + \sum_i^{H \times W}(y_{p}^i)^2 + \tau},
\end{align}
where $i$ denotes the $i$-th pixel of the feature map, $\tau$ is a smooth term to avoid zero division.

The design of  structure constraint module in teeth branch is the same as the dental plaque. And $L^t_{SCM}$ and $L^p_{SCM}$ have the same form of computation.

\subsection{Network Training}
The whole architecture can be trained in an end-to-end manner using the total loss function:
\begin{align}
& L = L_{S}^{p} + L_{S}^{t} + L_{SCM}^{p} + L_{SCM}^{t} + L_{CCM},
\end{align}
where $L_S$ is the standard implementations of the segmentation loss function for dental plaque or teeth instances, $L_{SCM}$ is the loss function for structure constraint module. $L_{CCM}$ is the loss for the contrastive constraint module.
Since boundary and mask are crossed linked by sharing one decoder, optimizing the loss for boundary can enhance the feature representation for mask predictions. Moreover,
optimizing the loss in contrastive constraint module can push confused regions belonging to different categories away from each other which benefit the mask prediction.

\section{The Stained Dental Plaque Dataset}

The lack of high-quality, open-source and finely annotated datasets is an important factor that has hindered the vigorous development of the field of dental plaque segmentation.
Our goal is to introduce a worthy benchmark with high-quality annotations to the community of dental plaque segmentation and improve the accuracy of computer-aided diagnosis.

\subsection{Dataset Collection and Annotation.}
Our datasets are collected by the dentists at Shanghai Shanda dental clinic.
According to the different oral digital imaging devices, i.e., 3-shape intraoral scanner and HD camera,  we put forward two datasets, which are referred as \textit{SDPSeg-S} (stained dental plaque segmentation dataset obtained by scanner) and \textit{SDPSeg-C} (stained dental plaque segmentation dataset obtained by camera), respectively.

The oral images in the \textit{SDPSeg-S} dataset are obtained by a high-resolution laser scanner (iTero, America). 3-shape intraoral scanner is an advanced oral digital impression system in the world which uses ultrafast optical cutting technology and confocal microscopy technology to capture multi frame 2D images to create 3D impressions in real time. 3-shape oral scan is comfortable, accurate and efficient, and will definitely become the trend of the future. Therefore, we collect 2D images of each patient's left view, right view, front view, maxillary and mandibular occlusal view generated by 3D oral scanner to construct the dataset.
The oral images in \textit{SDPSeg-C} dataset are taken by an HD camera (Cannon 750D, Japan) and also collect from each patient's left view, right view and front view, maxillary and mandibular occlusal view. This collection method is more convenient and is one of the main collection methods at present.

To get a more fine dental plaque contour, for both datasets, we first crop out every teeth visible in the field of vision, normalize its size, and then annotate the dental plaque and teeth area pixel by pixel using annotation tool. All annotations are performed by 6 people guided by professional dentists, and the annotation results are rechecked under strict quality control of the experienced dentist.

\subsection{Dataset Statistics}

\textbf{Dataset splitting.} Totally, we collect 565 and 1304 images for \textit{SDPSeg-S} and \textit{SDPSeg-C} datasets, respectively.
The two datasets are both randomly split into training set, test set and verification set with a ratio of 8:1:1, which contain 455, 57, 57 images and  1043, 131, 130 images, respectively.

	\begin{figure}[ht]
		\centering
		\centerline{\includegraphics[width=0.95\linewidth]{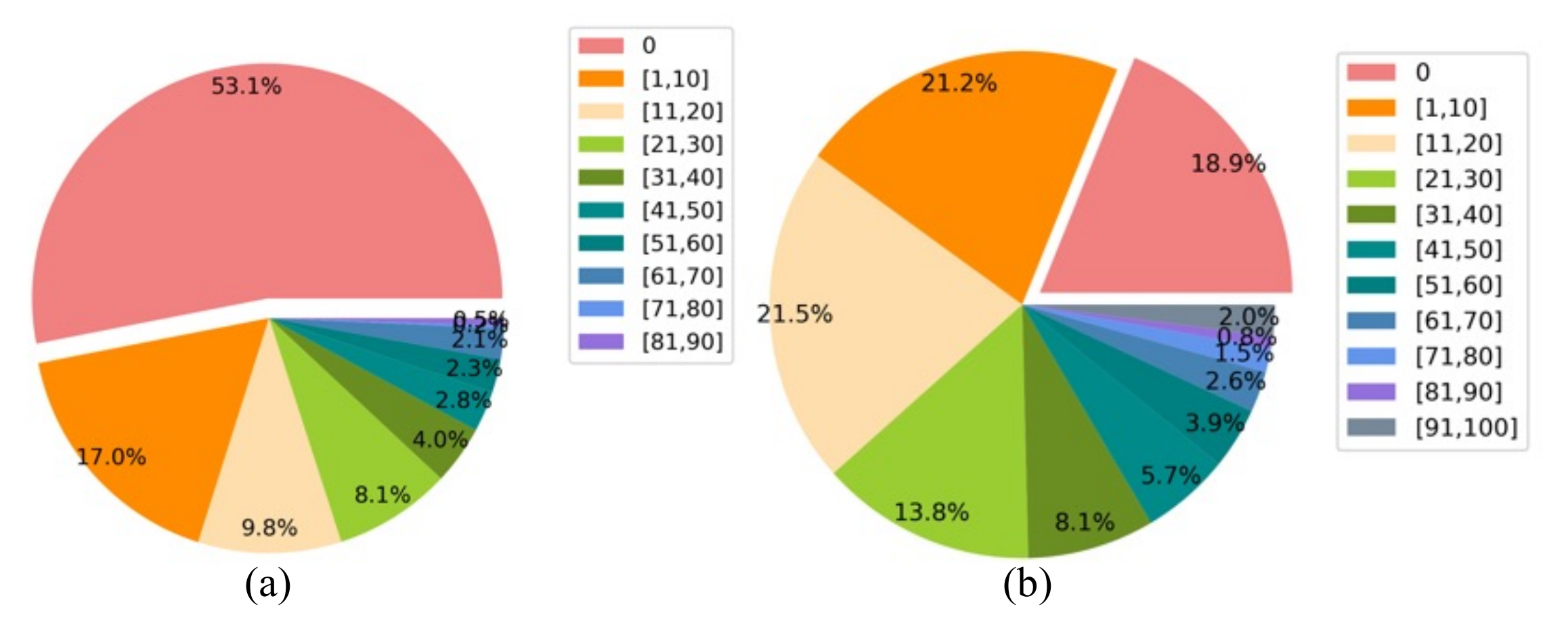}}
		\caption{The proportion of dental plaque. (a) and (b) show the distributions of the ratio of dental plaque to teeth on \textit{SDPSeg-S} and \textit{SDPSeg-C}, respectively.}
		\label{fig:data2}
	\end{figure}
	\begin{figure}[ht]
		\centering
		\centerline{\includegraphics[width=0.95\linewidth]{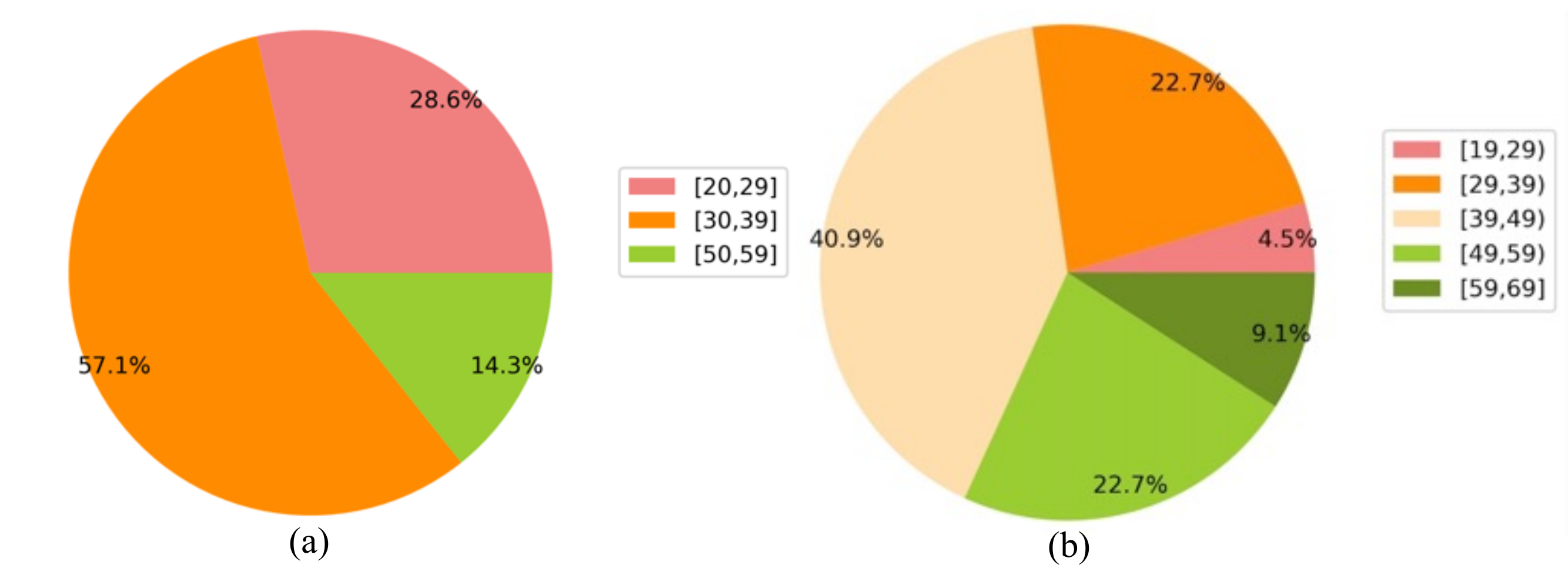}}
		\caption{The distributions of age. (a) and (b) show the distribution of age on \textit{SDPSeg-S} and \textit{SDPSeg-C}, respectively. Compared with the \textit{SDPSeg-S} dataset, the \textit{SDPSeg-C} dataset has a larger age span, including all ages from the adolescent to the elderly.}
		\label{fig:data1}
	\end{figure}
\textbf{The proportion of plaque.} Both of the two datasets contain three categories: teeth, dental plaque and background.
In this section we describe the complex distribution of dental plaque. The \textit{SDPSeg-S} and \textit{SDPSeg-C} are divided into 11 severity levels from 0$\%$, 1-10$\%$ to 91-100$\%$ by calculating the ratio of dental plaque area to teeth area in each oral image.
Fig. \ref{fig:data2} (a) shows the proportion of severity level on \textit{SDPSeg-S} dataset, where oral images with and without dental plaque account for 46.9 and 53.1 percent, respectively.
Fig. \ref{fig:data2} (b) shows the proportion of each severity level in \textit{SDPSeg-C} dataset, where oral images with and without denal plaque account for 81.1 and 18.9 percent, respectively.
There are significant differences in dental plaque distribution among different ages, oral sites and disease states. The oral images in \textit{SDPSeg-S} dataset \textit{SDPSeg-C} dataset were collected from people with different degrees of oral disease and age distribution, so the distribution of dental plaque varies greatly.

Observing from the statistical result of images with dental plaque, the data of mild dental plaque, i.e., 1-30$\%$, accounts for major on both \textit{SDPSeg-S} and \textit{SDPSeg-C}.
On \textit{SDPSeg-S} dataset , the proportion of dental plaque in all data is less than 90\%, while on the \textit{SDPSeg-C} dataset with oral problems, the data with dental plaque accounting for more than 90\% of teeth occupy 2\% of the dataset. Overall, the number of stained dental plaque data on both datasets conforms to the real situation of dental plaque distribution in clinical diagnosis.

\textbf{The distribution of age.} All oral data in the two datasets are both  provided by adults. Fig. \ref{fig:data1} illustrates the distribution of age on \textit{SDPSeg-S} and \textit{SDPSeg-C} datasets. \textit{SDPSeg-S} dataset covers three age groups: 20-29, 30-39, and 50-59 years old, among which the age group 30-39 years old accounts for the largest proportion, 57.1\%. \textit{SDPSeg-C} dataset includes the data of all ages from the adolescent to the elderly, among which the age group 29-39 accounts for 22.7\%, the age group 39-49 accounts for 40.9\%, and \textit{SDPSeg-C} dataset also contains stained dental plaque images of some older groups, i.e., the age group 59-69 accounts for 9.1\%.

\section{Experimental Results}
\subsection{Evaluation Metric}
\label{metric}
In this paper, we employ the commonly used evaluation metric MIoU and Dice in medical image segmentation. In addition, to provide doctors with a more objective reference for the area ratio of dental plaque to teeth in clinical medical diagnosis, we propose a new evaluation metric pixel ratio (PR), which is defined as follows:
\begin{align}
    & PR = \frac{\sum_{i=0,j=0}^{H\times W} Pi_{p}}{\sum_{i=0,j=0}^{H\times W}  {Pi_{t} + pi_{p}}}
\end{align}
Where $H$ and $W$ represent the height and width of the image, respectively.
$Pi_{p}$ represents the pixels belonging to the category of dental plaque, and $Pi_{t}$ represents the category of teeth. We define that when the value of $ \| PR_{gt} - PR_{pre}\|$ is less than or equal to 5\%, the segmentation result achieve the effect of assisting doctors to diagnose. So the accuracy rate of the ratio of dental plaque to teeth, PR\%, is defined as:
\begin{align}
    & PR \% = \frac{\sum { _{i=0}^M PR_{True}}} {\sum {_{i=0}^M PR_{True} + PR_{False}}}
\end{align}
Where $PR_{True}$ represent the values of $ \|PR_{gt} - PR_{pre}\|$ or $ \|PR_{gt} - PR_{cli}\|$ are less than or equal to 5\%, $PR_{False}$ represents the values are more than 5\%. $PR_{gt}$ is the PR of the groundtruth, $PR_{pre}$ is the PR of the prediction, and $PR_{cli}$ is the PR of the clinician's estimation.

\subsection{Implementation Details}
During the training stage, we use the data division method of the proposed dataset, and the image dimensions used for training is 128$\times$128 which is the original dimensions
of the curated dataset. We use Adam optimizer \cite{b55} with momentum $=0.9$, $\beta_1 = 0.9$, $\beta_2 = 0.99$ and $\epsilon = 10^{-8}$ to train our model. The batch size is set to 16 on both datasets. The total epochs is set to 120, and the initial learning rate is set to 1e-4 and decreased by multiplying 0.1 every 40 epochs on \textit{SDPSeg-S} dataset. And the total epochs is set to 300, and the initial learning rate is set to be 1e-4 and decreased by multiplying 0.1 every 100 epochs on \textit{SDPSeg-C} dataset. The hyper parameters $\alpha$, $\beta$ in compound loss function $L_{SCM}$ are set to 0.1, 1 on both two datasets.
We implemented our method using PyTorch with a single Tesla V100-SWM2 GPU.

	\begin{figure*}[ht]
		\centering
		\centerline{\includegraphics[width=1\linewidth]{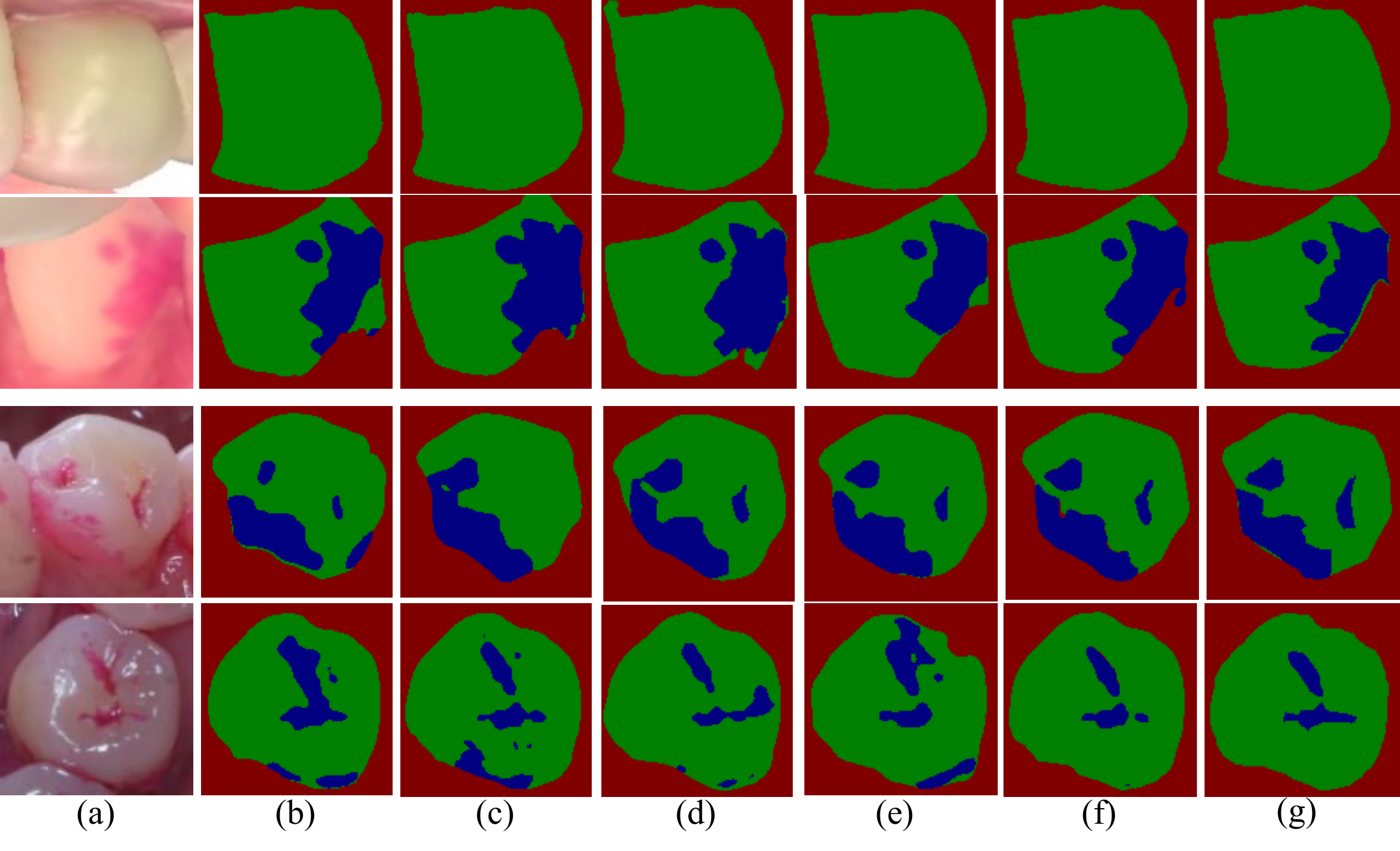}}
		\vspace{-6pt}
		\caption{Qualitative visual results on dental plaque segmentation task compared with state-of-the-art methods. The first two rows are the results on \textit{SDPSeg-S} dataset, and the last three rows are the results on \textit{SDPSeg-C} dataset. (a) are the original images; (b) are the results of UNet \cite{b12}; (c) are the results of Attention UNet \cite{b49}; (d) are the results of Channel UNet\cite{b51}; (e) are the results of TransUNet \cite{b53}; (f) are the results of SDNet (Ours); (g) are groundtruths. Our method can segment relatively complete dental plaque regions and has a stronger ability to recognize details.}
		\label{fig:vis1}
		\vspace{-6pt}
	\end{figure*}
	
\subsection{Results and Analysis}

\begin{table*}[!t]
	\begin{center}
		\caption{Performance of SDNet and other segmentation methods on \textit{SDPSeg-S} and \textit{SDPSeg-C} datasets.}		
		\label{table:1}
		\scalebox{1.1}{
			\begin{tabular}{cccccccccc}
				
				\hline
				\multicolumn{1}{c}{\multirow{3}*{Methods}} &
				\multicolumn{4}{c}{\textit{SDPSeg-S}} & \multicolumn{4}{c}{\textit{SDPSeg-C}} &
				\multicolumn{1}{c}{}\\
				\cline{2-10}
				\multicolumn{1}{c}{} &\multicolumn{2}{c}{Teeth} & \multicolumn{2}{c}{Plaque} & \multicolumn{2}{c}{Teeth} & \multicolumn{2}{c}{Plaque}&
				\multicolumn{1}{c}{\multirow{2}*{Time(ms)}}\\
				\cline{2-9}
				\multicolumn{1}{c}{}& MIoU & Dice & MIoU & Dice & MIoU & Dice & MIoU & Dice &\\
				\hline
				UNet \cite{b12} & 89.07 & 93.90 & 70.24 & 67.40 & 86.80 & 91.15 & 75.44 & 75.78 & 129.21\\
				ResUNet \cite{b48} & 86.38 & 93.08 & 67.33 & 65.72 & 86.00 & 90.04 & 72.54 & 70.94 & 144.59\\
				Attention UNet \cite{b49} & 89.61 & 94.46 & 70.66 & 68.09 & \textbf{87.56} & 91.01 & 76.75 & 76.45 & 157.94\\
				R2 UNet \cite{b50} & 71.31 & 86.88 & 51.39 & 53.97 & 79.96 & 87.18 & 64.10 & 62,59 & 183.12\\
				Channel UNet \cite{b51} & 89.76 & 94.35 & 68.24 & 63.26 & 87.33 & 90.98 & 75.97 & 75.56 & 155.69\\
				KiUNet \cite{b52} & 85.54 & 92.94 & 66.04 & 64.70 & 84.61 & 90.50 & 70.41 & 70.34 & 160.75\\
				TransUNet \cite{b53} & 89.08 & 94.30 & 71.78 & 70.65 & 87.38 & 91.42 & 75.95 & 76.05 & 236.57\\
				MedT \cite{b54} & 70.02 & 85.80 & 46.27 & 43.29 & 77.29 & 83.05 & 62.82 & 59.23 & 447.90\\
				SDNet (Ours) & \textbf{90.35} & \textbf{95.58} & \textbf{80.08} & \textbf{77.02} & 87.55 & \textbf{94.94} & \textbf{82.15} & \textbf{83.08} & \textbf{57.43}\\
	            \bottomrule	
		    \end{tabular}}
	\end{center}
\end{table*}


\textbf{Results on \textit{SDPSeg-S} Datasets.} Table \ref{table:1} shows a comparison of our method on \textit{SDPSeg-S} dataset with some recent methods including UNet and UNet variants in the field of medical segmentation.
To be fair, all the comparison methods in Table \ref{table:1} use the same training methods and training parameters as ours, and the data argument methods are the same (including horizontal flip and vertical flip), too.
SDNet outperforms all the methods on the this dataset and the improvement is particularly significant on the category of dental plaque. SDNet achieves a MIoU of 90.35, Dice of 95.58 on teeth category, and a MIoU of 80.08, Dice of 77.02 on dental plaque category. We can observe an improvement of 1.28\% and 1.68\% over UNet on the teeth category in MIoU and Dice, respectively, and the improvement is particularly large on the more complex dental plaque category, which could be achieved 9.84\% and 9.62\% on this category. Compared with TransUNet \cite{b53}, which takes advantage of both Transformers \cite{b56} and UNet, SDNet also outperforms it by 1.27\% in MIoU and 1.28\% in Dice on the teeth category, and 8.30\% and 6.37\% on the dental plaque category. As shown in the $1^{st}$ and $2^{nd}$ rows of Fig. \ref{fig:vis1}, we present the visual comparisons on \textit{SDPSeg-S}.
All of the above methods can obtain outstanding segmentation results for the simple segmentation samples containing only teeth category, e.g., the sample shown in the $1^{st}$ row of Fig. \ref{fig:vis1}. However, for the challenging segmentation samples with extensive and variform dental plaque, e.g., the samples shown in the $2^{nd}$ row of Fig. \ref{fig:vis1}, our method can obtain clear dental plaque contours and precise details compared with other methods. Therefore, we observe that the increase of MIoU and Dice on teeth category in Table \ref{table:1} is much smaller than that on the more complex category dental plaque.


\textbf{Results on \textit{SDPSeg-C} Dataset.} The quantitative results of our SDNet and other methods on \textit{SDPSeg-C} dataset are also shown in Table \ref{table:1}. 
Our SDNet almost outperforms the other methods on all metrics, obtains a MIoU of 87.55\% and a Dice of 94.94\% on the teeth category, and a MIoU of 82.15\% and a Dice of 83.08\% on the dental plaque category. The improvement in Dice can be 3.79\% and 7.30\% on the dental palque category compared with UNet.
Our SDNet also achieves 20.49\% improvement in Dice on dental plaque category as compared to R2UNet\cite{b50}, 23.85\% improvement in Dice as compared MTNet\cite{b54} which uses transformer mechanism to encode long-range dependencies and learn representations that are highly expressive. 
The visual comparisons on \textit{SDPSeg-C} dataset can be observed in Fig. \ref{fig:vis1} ($3^{rd}-5^{th}$ rows). The $3^{rd}$ row shows the case that 
the dental plaque occupies the whole teeth and has a smooth contour, almost all methods can get an excellent segmentation result. 
However, observing the $4^{th}$ and $5^{th}$ rows in Fig. \ref{fig:vis1}, when the distribution of dental plaque becomes more random, our segmentation results perform better than other methods, which produces prediction masks with nearly the same structures and shapes of dental plaque as compared to the groundtruth masks, especially on the areas with fine boundaries.

\textbf{Inference Time.} Our method also achieves the optimal result in inference time, which has a shorter prediction time for an input image, just 57.43ms, compared with other methods. We can observe that the inference time of our method is less than half compared with the simplest UNet, and the inference time of the more complex method MedT is nearly 8 times than our method.  
Note that, our proposed two constraint modules, i.e., contrastive constraint module and structural constraint module, are both utilized only in training stage to help two branches better focus on category-specific feature which can improve segmentation performance and do not introduce additional computation and parameters during inference. 

\textbf{Application in Computer Aided Diagnosis.} The ratio of dental plaque to teeth is an important reference for the dental plaque index to evaluate the severity of dental plaque. Table \ref{table:2} shows the accuracy of our method in predicting the ration of dental plaque to teeth compared with doctors $\textit{Dr.}$ and nurses $\textit{N.}$ (2 doctors and 2 nurses are from the Shanghai Shanda dental clinic).
PR\% is an index introduced in section \ref{metric} to evaluate the accuracy of the ratio of denal plaque to teeth. We find that the accuracy of evaluation varies obviously among different doctors and nurses because of the different experience levels of physicians. Our accuracy reaches 84.21\% on the \textit{SDPSeg-S} dataset, which is 12.28\% and 42.10\% higher than experienced doctor and nurse, respectively. And on the \textit{SDPSeg-C} dataset, our method also achieves optimal results, increasing 35.11\% and 36.64\% compared with the experienced doctor and nurse. The above results indicate that our method possesses promising application prospect in computer aided diagnosis.

\textbf{Generalizability experment on other datasets.} 
We conducted generalization experiments on the polyp segmentation datasets CVC-ClinicDB \cite{b61} and CVC-ColonDB \cite{b62} to verify the generalization of our method on other datasets. Table \ref{table:9} shows the performance of our method compared with other state-of-the-art methods. To keep the fairness of the experiments, all methods of the Table \ref{table:9} follow \cite{b63} advice and take exactly the same training and testing dataset division. Compared with UNet \cite{b12}, our method improves greatly in both datasets. On CVC-ClinicDB, the improvement of our method in MIOU is as high as 10.7\%, and on CVC-ColonDB, the improvement reaches 31.4\% and 29.1\% in MDice and MIoU. We can observe an improvement of 1.9\% and 0.8\% over the representative method PraNet \cite{b63} in MDice and MIoU on CVC-ClinicDB dataset, and the improvement is particularly large on the more difficult dataset CVC-ColonDB, which could be achieved 11.4\% and 9.5\% in MDice and MIoU. Compared with the latest method \cite{b64} for polyp segmentation, our method achieves similar results on CVC-ClinicDB dataset, and on CVC-ColonDB dataset, our method improves by7.3\% and 6.5\% on MDice and MIoU, respectively. In other segmentation task, our method can also achieve the same or much higher accuracy on the polyp segmentation datasets CVC-ClinicDB \cite{b61} and CVC-ColonDB \cite{b62} than state-of-the-art methods, indicating that our method is not only designed for dental plaque segmentation task, but also effective on other tasks.
	
\subsection{Ablation Study}

	\begin{figure}[ht]
		\centering
		\centerline{\includegraphics[width=1\linewidth]{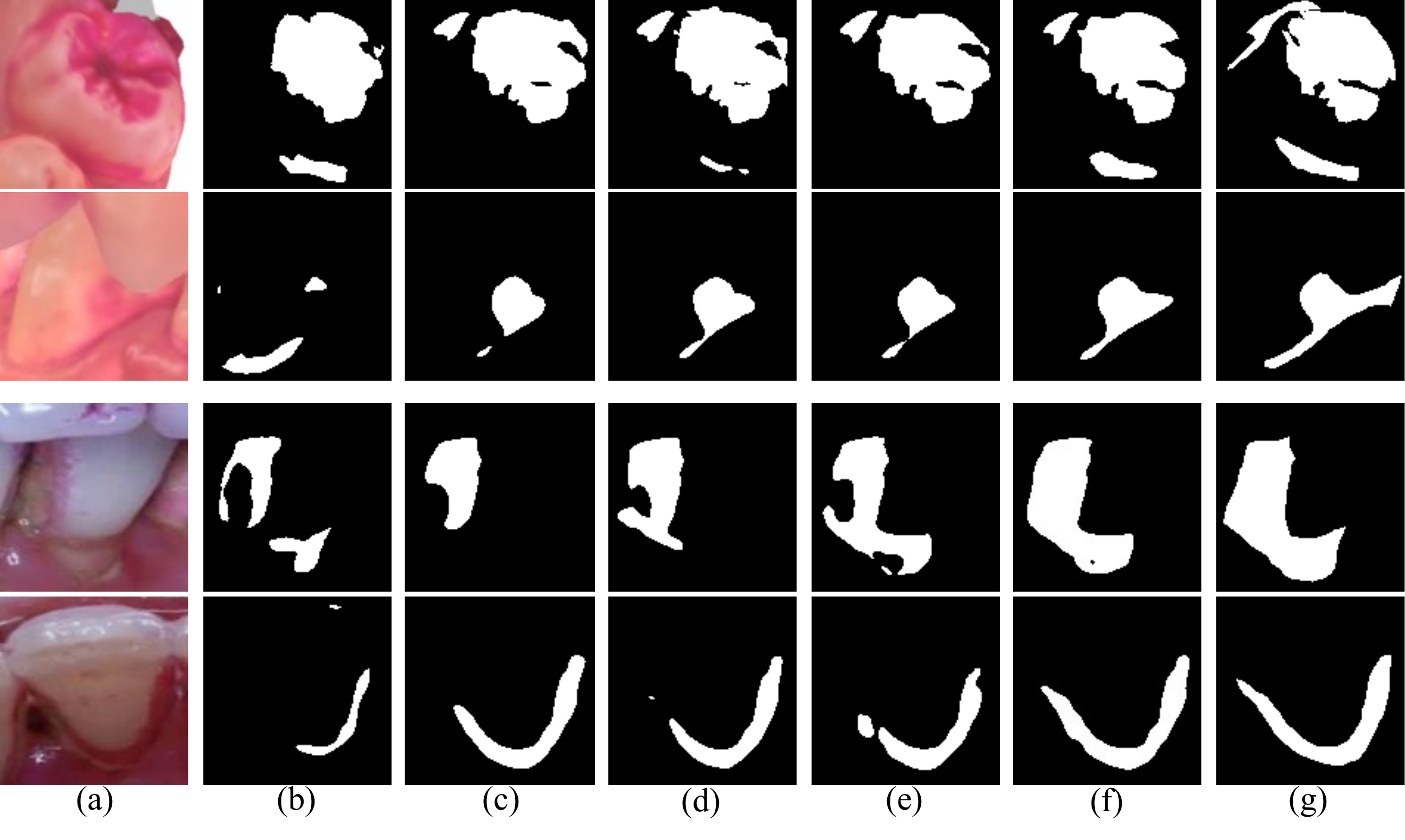}}
		\caption{Visual comparison of the results of dental plaque segmentation between different configuration: (a) is the original images; (b) is the baseline UNet; (c) is the SD structure; (d) is SD+CCM ; (e) SD+SCM; (f) is SDNet(Ours); (g) is groundtruths. Note that the effect of segmentation has been improved with the addition of each component. SD denotes the semantic decomposition structure in \ref{semantic decomposition}}
		\label{fig:vis2}
	\end{figure}

We conduct ablation studies to investigate three questions regarding our SDNet: 1) the contribution of each key designed component to our model performance;
2) the impact of different loss functions for constrain the boundary information introduced by SCM; 3) the impact of different integration positions for CCM.

\begin{table}[!t]
	\begin{center}
		\caption{The accuracy(PR\%)of our method in estimating the ratio of dental plaque to teeth compared with doctors and nurses on \textit{SDPSeg-S} dataset and \textit{SDPSeg-C} dataset.}		
		\label{table:2}
		\scalebox{0.95}{
			\begin{tabular}{cccccc}
				
				\hline
				\multicolumn{1}{c}{\multirow{1}*{Datasets}} &\multicolumn{1}{c}{$\textit {Dr.} 1$} & \multicolumn{1}{c}{$\textit {Dr.} 2$} &  \multicolumn{1}{c}{$\textit {N.} 1$} & \multicolumn{1}{c}{$\textit {N.} 2$} & \multicolumn{1}{c}{Ours} \\\hline
				\textit{SDPSeg-S} (PR\%) & 71.93 & 43.86 & 35.09 & 42.11 & \textbf{84.21} \\
				\textit{SDPSeg-C} (PR\%) & 41.22 & 40.45 & 25.95 & 39.69 & \textbf{76.33} \\
	            \bottomrule	
		\end{tabular}}
	\end{center}
\end{table}

\begin{table}[!t]
	\begin{center}
	    \vspace{-6pt}
		\caption{The Generalizability performance compared with other state-of-the-art polyp segmentation methods on CVC-ClinicDB and CVC-ColonDB datasets.}	
		\label{table:9}
		\scalebox{1.15}{
			\begin{tabular}{ccccc}
				\hline
				\multicolumn{1}{c}{\multirow{2}*{Models}} &\multicolumn{2}{c}{ClinicDB} & \multicolumn{2}{c}{ColonDB}  \\
				\multicolumn{1}{c}{}& MDice & MIoU & MDice & MIoU \\
				\hline
				UNet \cite{b12} & 0.823 & 0.750 & 0.512 & 0.444 \\
				ResUNet \cite{b50} & 0.779 & - & - & -\\
				PraNet \cite{b63} & 0.899 & 0.849 & 0.712 & 0.640\\
				SANet \cite{b64} & 0.916 & \textbf{0.859} & 0.753 & 0.670\\
				SDNet(ours) & \textbf{0.918} & 0.857 & \textbf{0.826} & \textbf{0.735}\\
	            \bottomrule	
		\end{tabular}}
	\end{center}
\end{table}

\begin{table}[!t]
	\begin{center}
		\caption{The contribution of main component in teeth category and dental plaque category on \textit{SDPSeg-S}.}		
		\label{table:3}
		\scalebox{0.82}{
			\begin{tabular}{ccccc}
				
				\hline
				\multicolumn{1}{c}{\multirow{2}*{Models}} &\multicolumn{2}{c}{Teeth} & \multicolumn{2}{c}{Plaque}  \\
				\multicolumn{1}{c}{}& MIoU & Dice & MIoU & Dice \\
				\hline
				UNet & 89.07 & 93.90 & 70.24 & 67.40 \\
				UNet + SD & 89.33 & 95.05 & 75.86 & 70.80 \\
				UNet + SD + SCM & 89.77 & 95.30 & 78.07 & 74.05\\
				UNet + SD + CCM & 89.92 & 95.37 & 77.34 & 73.05\\
				UNet + SD + SCM + CCM (SDNet)  & \textbf{90.35} & \textbf{95.58} & \textbf{80.08} & \textbf{77.02}\\
				
	            \bottomrule	
		\end{tabular}}
	\end{center}
\end{table}

\begin{table}[!t]
	\begin{center}
	    \vspace{-6pt}
		\caption{The contribution of main component in teeth category and dental plaque category on \textit{SDPSeg-C}.}		
		\label{table:4}
		\scalebox{0.82}{
			\begin{tabular}{ccccc}
				
				\hline
				\multicolumn{1}{c}{\multirow{2}*{Models}} &\multicolumn{2}{c}{Teeth} & \multicolumn{2}{c}{Plaque}  \\
				\multicolumn{1}{c}{}& MIoU & Dice & MIoU & Dice \\
				\hline
				UNet & 86.80 & 91.15 & 75.44 & 75.78 \\
				UNet + SD & 86.81 & 91.04 & 80.15 & 80.31 \\
				UNet + SD + SCM & 87.06 & 91.17 & 81.19 & 81.96\\
				UNet + SD + CCM & 86.83 & 90.95 & 81.85 & 82.68\\
				UNet + SD + SCM + CCM (SDNet) & \textbf{87.55} & \textbf{91.51} & \textbf{82.15} & \textbf{83.08}\\
				
	            \bottomrule	
		    \end{tabular}}
	\end{center}
\end{table}

\begin{figure}[ht]
	\centering
	\centerline{\includegraphics[width=1\linewidth]{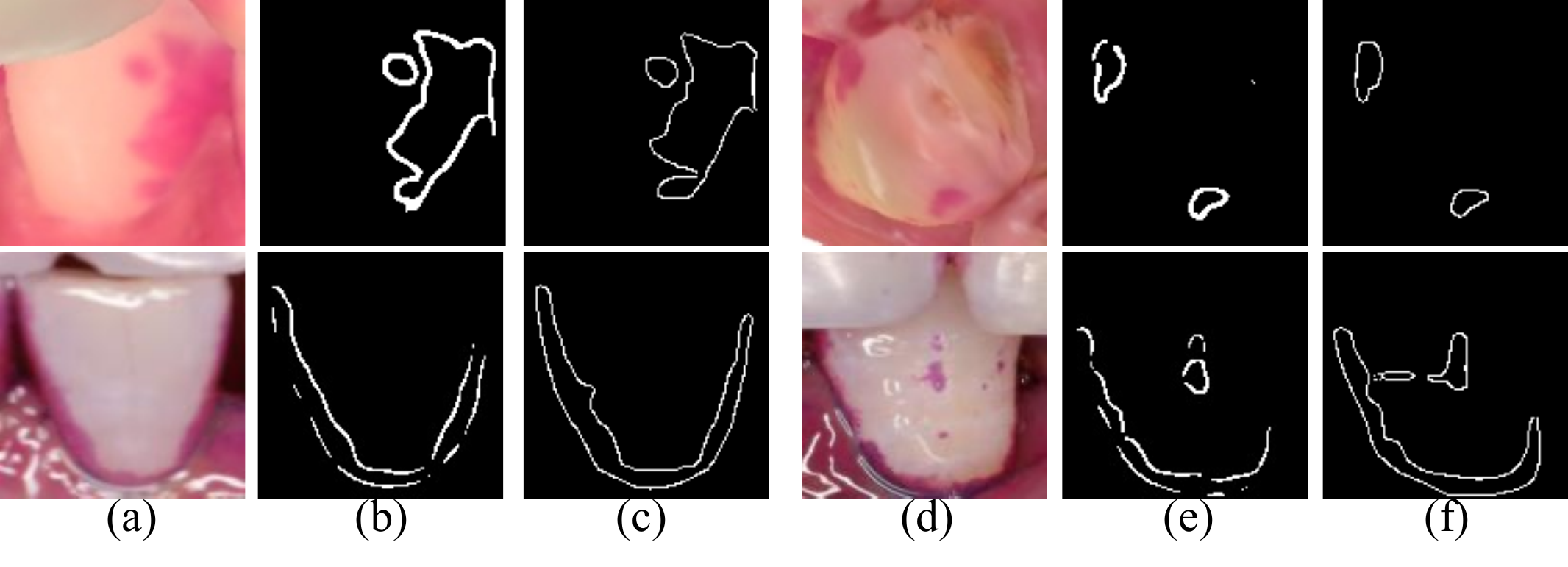}}
	\caption{Results of boundary prediction in auxiliary branch. The first row is the results of boundary prediction on \textit{SDPSeg-S}, and the second row is the result of boundary prediction on \textit{SDPSeg-C} dataset. (a),(d) are the original images; (b),(e) are boundary predictions results; (c),(f) are groundtruths. The boundary prediction results in the auxiliary branch are close to groundtruths, which can supplement the boundary information.}
	\label{fig:vis3}
\end{figure}
	
\textbf{Impact of each component.} Table \ref{table:3} and Table \ref{table:4} show the performance comparisons in terms of MIoU and Dice among different configurations, including semantic decomposition (SD) structure for decoupling dental plaque and teeth instances, contrastive constraint module (CCM) and structure constraint module (SCM), on \textit{SDPSeg-C} and \textit{SDPSeg-S} datasets, respectively. We investigate the impact of above three key components by gradually integrating each component into our framework.
(1) Baseline (UNet): the baseline starts from a UNet \cite{b12} network by feeding each oral image into it for teeth and dental plaque segmentation. 
(2) UNet + SD: the baseline is upgraded with a semantic decomposition (SD) structure. Compared with the baseline UNet, the performance is improved obviously on both two datasets, especially on the dental plaque category. 
(3) UNet + SD + SCM: the baseline is extended with SCM. By adding SCM on the setting (2), the performance gets a significant improvement on the results of dental plaque category compared with on the teeth category, which presents that our SCM can focus on the fine boundaries of dental plaque with various shapes and improve the dental plaque segmentation. 
(4) UNet + SD + CCM: the baseline is  extended with CCM. By simply adding CCM on the setting (2), the performance also gets an obvious improvement on the results of dental plaque category.
(5) UNet + SD + SCM + CCM: Compared with setting (2), we simultaneously add SCM and CCM to introduce stronger constraints to distinguish the features in the semantic-blur regions and provide complete structural information for dental plaque segmentation. Our final configuration gains best performance, which better demonstrates our effectiveness of category-specific semantic decomposition strategy and additional constraints.

\begin{table}[!t]
	\begin{center}
		\caption{The ablation of different loss functions introduced by SCM in teeth and dental plaque categories on \textit{SDPSeg-S} and \textit{SDPSeg-C} datasets in Dice(\%).}		
		\label{table:5}
		\scalebox{0.9}{
			\begin{tabular}{ccccccc}
				
				\hline
				\multicolumn{1}{c}{\multirow{2}*{SD + CCM}} &
				\multicolumn{1}{c}{\multirow{2}*{BCE Loss}} &
				\multicolumn{1}{c}{\multirow{2}*{Dice Loss}} &
				\multicolumn{2}{c}{\textit{SDPSeg-S}} & \multicolumn{2}{c}{\textit{SDPSeg-C}}\\
				\cline{4-7}
				\multicolumn{3}{c}{}& Teeth & Plaque & Teeth & Plaque\\
				\hline
				$\surd$ &  &  & 95.37 & 73.05 & 90.95 & 82.68\\
				$\surd$ & $\surd$ &  & 95.26 & 74.81 & 91.43 & 82.51\\
				$\surd$ &  & $\surd$ & 95.26  & 76.59 & 90.52& 81.49\\
				$\surd$ & $\surd$ & $\surd$ & \textbf{95.58} & \textbf{77.02} & \textbf{91.51} & \textbf{83.08}\\
	            \bottomrule	
		    \end{tabular}}
	\end{center}
\end{table}

\begin{table}[!t]
	\begin{center}
		\caption{The ablation of the position of CCM betweenthe teeth and dental plaque branchs on \textit{SDPSeg-S} and \textit{SDPSeg-C} datasets in Dice(\%).}		
		\label{table:6}
		\scalebox{0.8}{
			\begin{tabular}{ccccccccc}
				
				\hline
				\multicolumn{1}{c}{\multirow{2}*{}} &
				\multicolumn{2}{c}{After $\textit f1$} & \multicolumn{2}{c}{After $\textit f2$} & \multicolumn{2}{c}{After $\textit f3$} &
				\multicolumn{2}{c}{Ours}\\
				\cline{2-9}
				\multicolumn{1}{c}{}& Teeth & Plaque & Teeth & Plaque & Teeth & Plaque & Teeth & Plaque\\
				\hline
				\textit{SDPSeg-S} & 95.47 & 76.09 & 95.38 & 76.29 & 95.53 & 76.38 & \textbf{95.58} & \textbf{77.02}\\
				\textit{SDPSeg-C} &91.12 & 80.11 & \textbf{92.30} & 81.80 & 90.96 & 80.54 & 91.51 & \textbf{83.08}\\
	            \bottomrule	
		\end{tabular}}
	\end{center}
\end{table}

\begin{table*}[!t]
	\begin{center}
		\caption{The ablation of training hyper-parameters $\alpha$ and $\beta$ in Eq. 6 on \textit{SDPSeg-S} and \textit{SDPSeg-C} datasets in MIoU(\%).}		
		\label{table:8}
		\scalebox{1}{
			\begin{tabular}{ccccccccccccc}
				
				\hline
				\multicolumn{1}{c}{\multirow{2}*{}} &
				\multicolumn{2}{c}{$\alpha$=0.1, $\beta$=1 (ours)} & \multicolumn{2}{c}{$\alpha$=0.2, $\beta$=1} & \multicolumn{2}{c}{$\alpha$=0.4, $\beta$=1} &
				\multicolumn{2}{c}{$\alpha$=0.6, $\beta$=1} &
				\multicolumn{2}{c}{$\alpha$=0.8, $\beta$=1} &
				\multicolumn{2}{c}{$\alpha$=1, $\beta$=1}\\
				\cline{2-13}
				\multicolumn{1}{c}{}& Teeth & Plaque & Teeth & Plaque & Teeth & Plaque & Teeth & Plaque & Teeth & Plaque & Teeth & Plaque\\
				\hline
				\textit{SDPSeg-S} & \textbf{90.35} & \textbf{80.08} & 90.23 & 79.74 & 89.62 & 79.20 & 89.25 & 79.03 & 89.56 & 79.06 & 89.16 & 77.40\\
				\textit{SDPSeg-C} & 87.55 & \textbf{82.15} & \textbf{87.62} & 81.62 & 87.43 & 80.97 & 87.55 & 81.43 & 86.93 & 80.53 & 87.35 & 80.49\\
	            \bottomrule	
		\end{tabular}}
	\end{center}
\end{table*}

Fig.\ref{fig:vis2} further demonstrates the visual performances of our proposed method. As shown in the segmentation results, SD structure which decouples teeth category and dental plaque category into two layers can pay more attention to dental plaque regions than UNet, the dental plaque instances can be more detailed and complete after adding CCM which models the difference between the two branches and SCM which can enhance the supervision of entire instance structure. 
Fig. \ref{fig:vis3} shows the boundary prediction results in our SCM on \textit{SDPSeg-S} and \textit{SDPSeg-C} datasets. It can be observed that SCM get a relatively complete dental plaque contours, which provides complete structure information for segmenting the whole instance accurately.

\textbf{Impact of each loss function for SCM.} In order to verify the ability of the compound function to supervise the boundary information introduced by the SCM, 
we reduce one of the loss functions respectively to conduct experiments, i.e. only BCE loss or only Dice loss.
The quantitative results shown in Table \ref{table:5} indicate that when gradually removing each loss from SCM, it brings a performance degradation from  83.08 to 82.51 (removing Dice loss) and 81.49 (removing BCE loss) for dental plaque category on \textit{SDPSeg-C}. And the degradation trend is the same on \textit{SDPSeg-S}, a decline of 2.21\% and 0.43\% when removing Dice loss removing BCE loss on dental plaque category, respectively.
The experimental results prove that when using Dice loss and BCE loss together to supervise the boundary prediction in SCM, the performance achieves the best. It demonstrates that Dice loss can solve the class imbalance problem that can't be solved only by binary cross entropy loss for boundary prediction.

\textbf{Optimum position of CCM}. We also set several integration positions of CCM, including after the first layer of decoder (After $f1$), after the second layer of decoder (After $f2$) and after the third layer of decoder (After $f3$), to compare with our  integration position before the first layer of decoder. As shown in Table \ref{table:6}, the position we choose to insert CCM achieves the best performances on both datasets in dental plaque category. Although the Dice of the position After $\textit{f2}$ is higher than our method in teeth category on \textit{SDPSeg-C} dataset, the size of feature map increases with the upsampling operation in the decoder which leads to the exponentially increasing of computing cost. And more importantly, the dice on more complex category dental plaque is nowhere near as good as ours. Our method is the optimal choice of balancing computing cost and accuracy.

\textbf{Ablation Experiment on training hyper-parameters.} In this part, we conduct ablation experiments on hyper-parameters $\alpha$ and $\beta$ in Eq. 6 of Loss function in the training process. We set up the ablation experiment for the hyper-parameters according to the boundary perception ability of the loss function BCE loss and Dice loss, Dice Loss is more advantageous to solve the problem of category imbalance caused by the large difference between foreground and background pixels. Therefore, we fix the hyper-parameter $\beta$ in front of Dice Loss as 1, and transform $\alpha$ to verify the influence of different hyper-parameter ratios on the loss function. As can be seen in Table \ref{table:8}, in the two datasets, the accuracy of both teeth category and dental plaque category generally presents a declining trend with the increase of $\alpha$. So we set $\alpha$ as 0.1 and $\beta$ as 1.

\section{CONCLUSION}
In this paper, we propose a semantic decomposition network (SDNet) which employs divide-and-conquer strategy for the task of dental plaque segmentation. Specifically, we employ semantic decomposition structure to focus on category-specific features in each branch, and then propose a contrastive constraint module (CCM) to maximize the distance between different category representations, a structural constraint module (SCM) to provide complete structure information for dental plaques with various shapes.
Besides, We construct a large-scale open-source dataset \textit{SDPSeg} consisting of stained dental plaque images for research and assessment of dental plaque segmentation.
Extensive experiments demonstrate the effectiveness of our proposed method.
More importantly, our method is more accurate than experienced doctors in estimating the ratio of dental plaque to teeth, which indicate that our method possesses promising application prospect in computer-aided diagnosis.

\end{document}